# Hybrid Car-Following Strategy Based on Deep Deterministic Policy Gradient and Cooperative Adaptive Cruise Control

Ruidong Yan, Rui Jiang, Bin Jia, Jin Huang, and Diange Yang

*Abstract*—Deep deterministic policy gradient (DDPG)-based car-following strategy can break through the constraints of the differential equation model due to the ability of exploration on complex environments. However, the car-following performance of DDPG is usually degraded by unreasonable reward function design, insufficient training, and low sampling efficiency. In order to solve this kind of problem, a hybrid car-following strategy based on DDPG and cooperative adaptive cruise control (CACC) is proposed. First, the car-following process is modeled as the Markov decision process to calculate CACC and DDPG simultaneously at each frame. Given a current state, two actions are obtained from CACC and DDPG, respectively. Then, an optimal action, corresponding to the one offering a larger reward, is chosen as the output of the hybrid strategy. Meanwhile, a rule is designed to ensure that the change rate of acceleration is smaller than the desired value. Therefore, the proposed strategy not only guarantees the basic performance of car-following through CACC but also makes full use of the advantages of exploration on complex environments via DDPG. Finally, simulation results show that the car-following performance of the proposed strategy is improved compared with that of DDPG and CACC.

*Note to Practitioners*—This article presents a new car-following strategy, which avoids the impact of deep deterministic policy gradient (DDPG) performance degradation on the system. In the proposed strategy, DDPG is replaced with cooperative adaptive cruise control (CACC) when the performance of DDPG is worse than that of CACC. Meanwhile, a switching rule is designed to guarantee that the change rate of acceleration is smaller than the threshold. Simulation results show that the performance of hybrid car-following strategy has been improved compared with that of only using CACC or DDPG. Moreover, the proposed strategy has the advantages of low computational burden, high real-time performance, and good scalability.

Manuscript received May 4, 2021; accepted July 24, 2021. This article was recommended for publication by Lead Guest Editor A. Si and Editor H. Gao upon evaluation of the reviewers' comments. This work was supported in part by the Fundamental Research Funds for the Central Universities under Grant 2021RC224 and in part by the National Natural Science Foundation of China under Grant 71971015. *(Corresponding authors: Bin Jia; Diange Yang.)*

Ruidong Yan, Rui Jiang, and Bin Jia are with the School of Traffic and Transportation, Beijing Jiaotong University, Beijing 100044, China (e-mail: bjia@bjtu.edu.cn).

Jin Huang and Diange Yang are with the School of Vehicle and Mobility, Tsinghua University, Beijing 100084, China (e-mail: ydg@tsinghua.edu.cn).

Color versions of one or more figures in this article are available at https://doi.org/10.1109/TASE.2021.3100709.

Digital Object Identifier 10.1109/TASE.2021.3100709

## I. INTRODUCTION

CAR-FOLLOWING is one of the fundamental functions of autonomous driving. A poor car-following performance will lead to congestion and traffic oscillation, thus wasting commuter's time and increasing energy consumption and pollution [1]–[3]. Thus, the car-following research of autonomous driving has attracted great attention [4]–[7]. The car-following strategy based on differential equation model has been widely used due to its good interpretation, such as the conventional cruise control [5], adaptive cruise control [6], and cooperative adaptive cruise control (CACC) [7]. As we all know, the real traffic environment is full of complexity and randomness [9], [10]. However, the performance of car-following strategy based on differential equation model will be decreased in real traffic environment due to the restriction of model itself.

Recently, the car-following strategy based on learning approach has been paid much attention due to the advantage of exploration on complex and unknown state space. A car-following model using multilayer feedforward neural network was designed [11]. Then, the fuzzy logic was introduced into the neural network-based car-following strategy [12]. It should be noticed that the car-following process was modeled as the Markov decision process (MDP) [13]. Therefore, reinforcement learning has been applied to achieve better car-following performance [14]–[17]. The neural network fitting $Q$-learning algorithm was applied to the car-following behavior with the camera original vision [14]. After that, a deep $Q$-network was applied to car-following by adding experience replay so that the training time and stability were improved [15]. Compared with deep $Q$-network, the deep deterministic policy gradient (DDPG) is more suitable for continuous action space. By considering the control delay and actual vehicle dynamics, DDPG could obtain the car-following performance close to the dynamic programming [16]. By investigating DDPG and model predictive control on whether considering the modeling error or not, it shows that DDPG has more advantages in the presence of uncertainties [17]. However, the car-following performance of DDPG is usually degraded by unreasonable reward function design, insufficient training, and low sampling efficiency.

As we all know, the differential equation model can interpret the vehicle motion and it is very helpful in many scenarios. However, this type of model-based car-following strategy is also restricted by the model itself. The car-following strategy using DDPG shows advantages in complex and unknown traffic environment due to the good ability of exploration on state space. However, the car-following performance of DDPG may be degraded by the above-mentioned problems. Obviously, both differential equation model and reinforcement learning have their own limitations. Thus, researchers try to combine them to achieve better car-following performance. A supervised learning method was proposed to improve the success rate of the training process, where the action was updated by a type of soft way between action of actor–critic learning and the one of model-based approach [18]. A semirule-based decision-making strategy was designed for heavy intelligent vehicles, where certain rules were applied to the reward function to make the strategy interpretable [19]. When the distance between two vehicles is smaller than the safe threshold, the output was set to be the minimum one to avoid collision [20]. However, the above literature could not make full use of advantages of differential equation model and reinforcement learning methods in the whole state space.

To address the above-mentioned problems, this study proposes a type of hybrid car-following strategy (HCFS) based on DDPG and CACC, which is inspired by [21] in the field of highway exiting planner to some extent. Different from [21], the action of the proposed strategy will be calculated at each frame in real time rather than one motion period spending 0.75 s. The main contribution of this study is as follows.

1) A type of HCFS is proposed to improve the car-following performance in the whole state space. The calculation of DDPG and CACC is independent at each frame under the MDP framework. Given a current state, two actions can be obtained from CACC and DDPG. The optimal action is corresponding to the one which offers a larger reward.
2) On the one hand, CACC is used to guarantee the basic car-following function when the performance of DDPG is poor. On the other hand, the proposed strategy makes full use of the exploration ability of DDPG when the state is beyond the limitation of differentiation equation model. Thus, the performance of HCFS is improved in the whole state space.
3) By considering the soft-switching mechanism, a rule is designed to guarantee the change rate of acceleration satisfying the constraint in real traffic environment.

The rest of this article is organized as follows. The MDP framework is described in Section II. The HCFS is presented in Section III. The simulation verification is presented in Section IV, and finally, conclusion is given in Section V.

## II. MDP FRAMEWORK

In this section, an overview is given in Section II-A. The state space, action space, and reward function are presented in Sections II-B–II-D, respectively.

### A. Overview

In this study, the car-following scenario is shown in Fig. 1. The vehicles are supposed to be driving on the same lane without lane changing. One leading vehicle and several following vehicles are chosen to form a platoon. Each following vehicle is manipulated by a distributed car-following strategy proposed in this study under the MDP framework. The initial state of the following vehicle is chosen as $s_k \in S$. Then, an action $a_k \in A$ is chosen according to the state–action mapping policy $\mu(a_k|s_k)$. By performing the action, the vehicle reaches the next state $s_{k+1}$ and a reward $r_k(s_{k+1}|a_k, s_k) \in R$ is obtained from the environment. There is an optimal policy $\mu^*(a_k|s_k)$ to get the maximum value of $E \sum \gamma^k r_k(s_{k+1}|a_k, s_k)$, which is the expectation of expected cumulative discounted reward within one episode. The goal of reinforcement learning is to find out such $\mu^*(a_k|s_k)$. The longitudinal motion of the following vehicle is under the MDP framework. Next, we introduce state space, action space, and reward function.

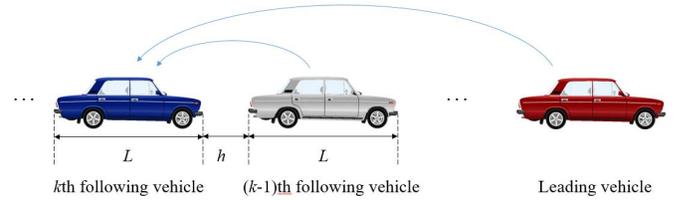

Fig. 1. Car-following scenario.

### B. State Space

The position $x_k$ and velocity $v_k$ of the ego-vehicle are supposed to be measured by the fusion of inertial navigation and GPS. The velocity of leading vehicle $v_0$ and the position of previous vehicle $x_{k-1}$ are supposed to be obtained by the V2X communication. To ensure safety, the distance between two adjacent vehicles should be greater than vehicle length $L$ plus safe space headway $h$. The distance deviation between the $k$th and $(k-1)$th vehicles is described as

$$e_{x\_k\_k-1} = x_k - x_{k-1} - (L + h). \tag{1}$$

The velocity error between the $k$th and $(k-1)$th vehicles is described as

$$e_{v\_k\_k-1} = v_k - v_{k-1}. \tag{2}$$

As we all know, the distance deviation, velocity error, velocity, and acceleration are very important in the car-following system [22]. Therefore, the state space of MDP should include the following features:

$$[e_{x\_k\_k-1}, e_{v\_k\_k-1}, e_{x\_k\_0}, e_{v\_k\_0}, v_k, a_k] \tag{3}$$

where $e_{v\_k\_0}$ and $e_{x\_k\_0}$ are velocity error and distance deviation between $k$th vehicle and leading vehicle, respectively.

### C. Action Space

All vehicles are supposed to be driving on the same lane without considering lane-changing behavior in this study. For the $k$th vehicle, the action is chosen as the target acceleration signal $a_k$. After receiving $a_k$, the bottom controller will drive the $k$th vehicle with this acceleration.

## *D. Reward Function*

As we all know, different designs of reward function will affect the car-following performance of DDPG. The reward function should be designed appropriately according to the actual situation. In this study, all vehicles are considered as a whole, which is kind of like a "small train." The smaller the velocity error among vehicles in the platoon is, the better the uniform stability of "small train" is. Therefore, the velocity error between the leading vehicle and the $k$th vehicle, i.e., ego-vehicle, should be considered in the reward function

$$r_{v\_k\_0} = -\omega_1 \frac{|e_{v\_k\_0}|}{v_{\max}} \quad (4)$$

where $v_{\max}$ is the maximum velocity, $|\cdot|$ represents the absolute value of a variable, and $\omega_1$ is a positive coefficient.

Meanwhile, driving comfort should also be considered. Similar to [20], the change rate of acceleration, i.e., jerk, is considered in the reward function

$$r_{\text{jerk}} = -\omega_2 \frac{|\text{jerk}|}{2a_{\max}/\Delta T}, \quad \text{jerk} = (a_k - a_{k-1})/\Delta T \quad (5)$$

where $a_{\max}$ and $\Delta T$ are maximum acceleration and time step, respectively, and $\omega_2$ is a positive coefficient.

By considering the above factors, the reward function is designed as

$$r = r_{v\_k\_0} + r_{\text{jerk}}. \quad (6)$$

## III. HCFS

In this section, a review of car-following strategy based on DDPG is presented in Section III-A. A review of car-following strategy based on CACC is given in Section III-B. Finally, HCFS is designed in Section III-C.

### *A. Review of Car-Following Strategy Based on DDPG*

DDPG is based on the MDP framework presented in Section II. The DDPG algorithm used in this study is the same as the algorithm in [23]. Since it is able to deal with multidimensional input and continuous action output with a relatively moderate calculation burden, DDPG has been applied to the car-following system [16], [17].

DDPG is one of the actor–critic algorithms. There are two current deep neural networks: actor network $\mu(s|\theta^\mu)$ and critic network $Q(s,a|\theta^Q)$, where $\mu$ is the state–action mapping policy, $Q$ is the $Q$ value function, and $\theta^\mu$ and $\theta^Q$ are net weight parameters. For the convergence of deep neural networks, two target networks $\mu'(s|\theta^{\mu'})$ and $Q'(s,a|\theta^{Q'})$ are also used with net weight parameters $\theta^{\mu'}$ and $\theta^{Q'}$. The action $a_k = \mu(s_k|\theta^\mu) + N_k$ is selected by current policy $\mu$ and exploration noise $N_k$. The current critic network is updated by minimizing root-mean-squared loss using gradient descent as

$$L_k = r(s_k, a_k) + \gamma \, Q'(s_{k+1}, \mu'(s_{k+1}|\theta^{\mu'})|\theta^{Q'}) - Q(s_k, a_k|\theta^Q). \quad (7)$$

The current actor network is updated by the sampled gradient as follows:

$$\nabla_{\theta^\mu} \mu|_{s_k} \approx \frac{1}{m} \sum_{k=1}^{m} \nabla_a Q(s, a|\theta^Q)|_{s=s_k, a=\mu(s_k)} \nabla_{\theta^\mu} \mu(s|\theta^\mu)|_{s_k}. \quad (8)$$

The target networks are updated by

$$\theta^{Q'} \leftarrow \tau \theta^Q + (1-\tau)\theta^{Q'} \quad (9)$$
$$\theta^{\mu'} \leftarrow \tau \theta^\mu + (1-\tau)\theta^{\mu'}. \quad (10)$$

*Remark 1:* According to [17], the performance of deep reinforcement learning is better than the conventional CACC based on differential equation model in a complex environment. Among deep reinforcement learning algorithms, DDPG not only has excellent continuous motion processing ability but also has a relatively moderate calculation burden. A good performance of DDPG is determined by a reasonable reward function design, sufficient training of the deep neural network, and high sample efficiency to a great extent. However, it is impossible to exhaust all traffic scenarios for pretraining of DDPG and difficult to design a suitable reward function for the whole state space. Moreover, the sampling efficiency is also very important. These difficulties will degrade the performance of car-following strategy based on DDPG in real traffic.

### *B. Review of Car-Following Strategy Based on CACC*

The typical CACC algorithm is based on differential equation. The highway capacity and traffic flow stability can be improved significantly through CACC as compared with ACC, which has been verified by the California PATH program [4]. Thus, a typical CACC similar to [4] and [5] is used in this study and the target acceleration command of $k$th following vehicle is presented as:

$$a_k = v_{k\_\text{last}} + k_1 e_{x\_k\_k-1} + k_2 e_{v\_k\_k-1} + k_3 e_{x\_k\_0} + k_4 e_{v\_k\_0} \quad (11)$$

where $v_{k\_\text{last}}$ is the velocity of the $k$th vehicle at the last frame, $e_{v\_k\_k-1}$ is the velocity error between $k$th and $(k-1)$th vehicles, $e_{v\_k\_0}$ is the velocity error between the $k$th vehicle and the leading one, and $k_1$–$k_4$ are positive coefficients.

*Remark 2:* Although a good performance of CACC could be achieved through differential equation model in most scenarios, the performance of CACC is limited by the model itself. When the real traffic environment is beyond the description of the model, the performance of CACC is degraded and even the traffic safety is threatened. Therefore, it is necessary to improve the performance of CACC based on differential equation model under complex and unknown environments.

### *C. HCFS*

In order to deal with the problems mentioned in Remarks 1 and 2, a type of HCFS is designed in this section.

HCFS is based on the MDP framework. As can be seen from Fig. 2, the calculation of DDPG and CACC is independent at each frame. First, given a current state $s_k$, two actions $a_\text{DDPG}$ and $a_\text{CACC}$ are calculated by DDPG and CACC, respectively.

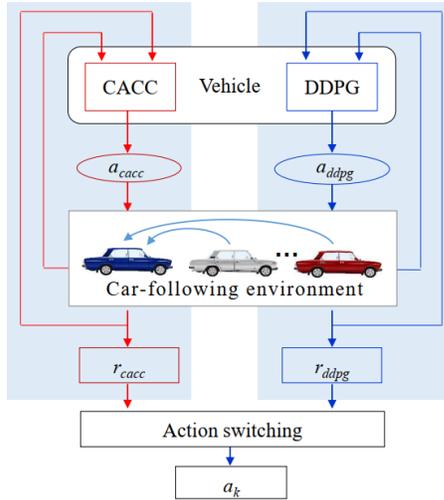

Fig. 2. HCFS.

After that, the following rule is used to select the action of HCFS:

$$a_k = \begin{cases} a_{\text{DDPG}}, & r_{\text{DDPG}} > r_{\text{CACC}} \\ a_{\text{CACC}}, & \text{else} \end{cases} \quad (12)$$

where $r_{\text{DDPG}}$ and $r_{\text{CACC}}$ represent the reward of DDPG and CACC, respectively, and $a_{\text{DDPG}}$ and $a_{\text{CACC}}$ represent the action of DDPG and CACC, respectively.

It should be noticed that the key of HCFS is the action switching between DDPG and CACC, which is based on the value of reward at each frame. In order to reduce the effect of perturbation caused by switching, a relatively soft-switching mechanism is introduced into (12) as

$$a_k = \begin{cases} (1-\beta)a_{\text{DDPG}} + \beta a_{\text{CACC}}, & r_{\text{DDPG}} > r_{\text{CACC}} \\ (1-\beta)a_{\text{CACC}} + \beta a_{\text{DDPG}}, & \text{else} \end{cases} \quad (13)$$

where $0 \leq \beta \leq 1$ is a coefficient.

Moreover, the change rate of acceleration through HCFS should satisfy constraint $|a_k - a_{k-1}| < \text{jerk}\,\Delta T$ at the $k$th and $(k-1)$th frames. When there is no switching, let $\alpha = 0$. When the switching occurs, let $\alpha = 1$. To satisfy the above constraint, the value of $\beta$ is chosen as

$$\beta = \begin{cases} 0, & \alpha = 0 \\ 0.5, & \alpha = 1. \end{cases} \quad (14)$$

To ensure that the choosing value of $\beta$ in the switching rule in the form of (13) satisfies the constraint, the following theorem is presented.

*Theorem:* Given an assumption that $a_1$ and $a_2$ satisfy constraints $|a_{1\_k} - a_{1\_k-1}| < \text{jerk}\,\Delta T$ and $|a_{2\_k} - a_{2\_k-1}| < \text{jerk}\,\Delta T$ respectively, when the acceleration is changed from $a_{1\_k-1}/a_{2\_k-1}$ to the one calculated by $a_{1k}$ and $a_{2k}$ according to the rule in the form of (13) and (14), the change rate is smaller than jerk under the framework of HCFS.

*Proof:* When there is no switching, the acceleration $a_1/a_2$ is changed from $a_{1\_k-1}/a_{2\_k-1}$ to $a_{1\_k}/a_{2\_k}$ according to the rule in the form of (13) and (14). Meanwhile, $|a_{1\_k} - a_{1\_k-1}| < \text{jerk}\,\Delta T$ and $|a_{2\_k} - a_{2\_k-1}| < \text{jerk}\,\Delta T$ can be obtained according to the assumption so that the change rate is smaller than jerk.

When the switching occurs, the acceleration is changed from $a_{1\_k-1}/a_{2\_k-1}$ to $0.5(a_{1k} + a_{2k})$ according to (13) and (14). Thus, the acceleration error between the $(k-1)$th and $k$th frames becomes the following equations:

$$0.5(a_{1\_k} + a_{2\_k}) - a_{1\_k-1} \quad (15)$$
$$0.5(a_{1\_k} + a_{2\_k}) - a_{2\_k-1}. \quad (16)$$

Under the framework of HCFS, the value of $a_{2\_k-1}$ is the same as $a_{1\_k-1}$ at the $(k-1)$th frame, i.e., $a_{1\_k-1} = a_{2\_k-1}$. Hence, the values of (15) and (16) are the same. Meanwhile, $a_{1\_k}$ and $a_{2\_k}$ are calculated independently of $a_{1\_k-1}$ and $a_{2\_k-1}$, respectively. According to the assumption, it yields

$$a_{1\_k-1} - \text{jerk}\,\Delta T < a_{1\_k} < a_{1\_k-1} + \text{jerk}\,\Delta T \quad (17)$$
$$a_{2\_k-1} - \text{jerk}\,\Delta T < a_{2\_k} < a_{2\_k-1} + \text{jerk}\,\Delta T. \quad (18)$$

Substituting $a_{1\_k-1} = a_{2\_k-1}$ into (18), we obtain

$$a_{1\_k-1} - \text{jerk}\,\Delta T < a_{2\_k} < a_{1\_k-1} + \text{jerk}\,\Delta T. \quad (19)$$

Substituting (17) and (19) into (15), it yields

$$-\text{jerk}\,\Delta T < 0.5(a_{1\_k} + a_{2\_k}) - a_{1\_k-1} < \text{jerk}\,\Delta T. \quad (20)$$

Obviously, (20) can be rewritten as

$$|0.5(a_{1\_k} + a_{2\_k}) - a_{1\_k-1}| < \text{jerk}\,\Delta T. \quad (21)$$

According to (15) and (21), the change rate of acceleration between the $(k-1)$th and $k$th frames is smaller than jerk when there is switching under the framework of HCFS. The proof is completed.

*Remark 3:* On the one hand, the car-following performance of DDPG is degraded by unreasonable reward function design, insufficient training, and low sampling efficiency presented in Remark 1. Thus, CACC is used to guarantee the basic car-following function when the performance of DDPG is poor. On the other hand, HCFS can make full use of exploration on complex state space of DDPG to deal with the cases beyond the limitation of differentiation equation model presented in Remark 2. Therefore, the car-following performance of HCFS will be improved in the whole state space.

## IV. SIMULATION VERIFICATION

In order to verify the effectiveness of the proposed scheme, comparisons among CACC, DDPG, and HCFS are carried out. As shown in Fig. 3, the velocity twice as slow of real vehicle experiment is used in this study [24], where the time step selected is 0.2 s.

The training environment for DDPG is chosen as a platoon, including one leading vehicle and six following vehicles. The ego-vehicle receives the position and velocity of previous and leading vehicles through the V2V communication. Meanwhile, the time delay between two adjacent vehicles through V2V is 5 ms, which is inevitable in real traffic. For the leading vehicle, velocity (i.e., training data) is chosen as the velocity from 600 s to the end in Fig. 3, acceleration is calculated by $a_k = (v_k - v_{k-1})/\Delta T$, and position is calculated by $x_{k+1} = x_k + v_k \Delta T + 0.5 a_k \Delta T^2$. For the $k$th following vehicle, acceleration $a_k$ is calculated by DDPG under the MDP framework constructed in Section II, velocity is calculated by $v_{k+1} = v_k + a_k \Delta T$, and position is calculated by $x_{k+1} = x_k + v_k \Delta T + 0.5 a_k \Delta T^2 + \eta$, where $\eta$ is the position deviation caused by time delay.

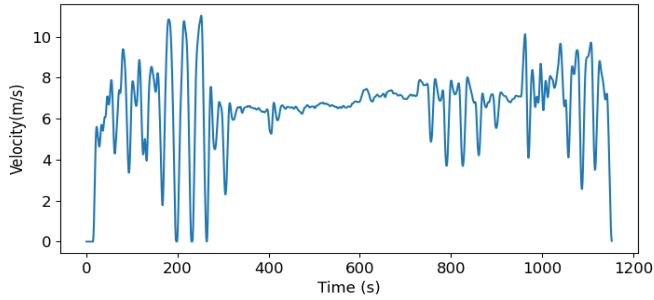

Fig. 3. Velocity of vehicle experiment.

TABLE I
PARAMETERS OF CACC

| Parameter | $k_1$ | $k_2$ | $k_3$ | $k_4$ |
|---|---|---|---|---|
| Value | 0.01 | 0.01 | 0.02 | 0.9 |

TABLE II
PARAMETERS OF DDPG

| Parameter | Value |
|---|---|
| Actor learning rate | 0.0001 |
| Critic learning rate | 0.001 |
| Experience replay buffer size | 500000 |
| Batch size | 32 |
| $\tau$ | 0.001 |
| $\gamma$ | 0.99 |

TABLE III
PARAMETERS OF HCFS

| Parameter | Value |
|---|---|
| $v_{\max}$ (m/s) | 100/3.6 |
| $a_{\max}$ (m/s$^2$) | 3 |
| $\Delta T$ (s) | 0.2 |
| $\omega_1$ | 10 |
| $\omega_2$ | 0.1 |

Parameters of CACC, DDPG, and HCFS are listed in Tables I–III, respectively.

### A. Case 1

In case 1, one leading vehicle and eight following vehicles are considered. The velocity between 200 and 220 s in Fig. 3 is chosen as the velocity of leading vehicle, which is the testing data in case 1. The testing data are not included in the training data. For each following vehicle, acceleration is calculated by CACC, DDPG, or HCFS, velocity is calculated by $v_{k+1} = v_k + a_k \Delta T$, and position is calculated by $x_{k+1} = x_k + v_k \Delta T + 0.5 a_k \Delta T^2 + \eta$. The time delay between two adjacent vehicles through V2V is 5 ms.

The green solid line in Fig. 4 is the velocity of leading vehicle, which is 0.37 m/s at the beginning. The velocity is rising quickly and then speed up to 10.76 m/s. Fig. 4 also shows the velocity of second, fourth, sixth, and eighth following vehicles. The velocity of following vehicles using CACC has periodic oscillation. The velocity of following vehicles using DDPG deviates from that of the leading vehicle between 2 and 8 s to some extent, while there is a zigzag oscillation between 12 and 20 s. Compared with CACC and DDPG, the vehicle velocity through HCFS is much more smooth during the entire process. Meanwhile, Table IV shows that the sum of absolute value of velocity error between leading and following vehicles using HCFS is 25.26 m/s, which is much smaller than that of CACC (159.39 m/s) and DDPG (146.22 m/s). The sum of the standard deviation of such velocity error using HCFS (0.04 m/s) is also smaller than that of CACC (0.22 m/s) and DDPG (0.16 m/s). Obviously, HCFS offers better performance than CACC and DDPG in case 1.

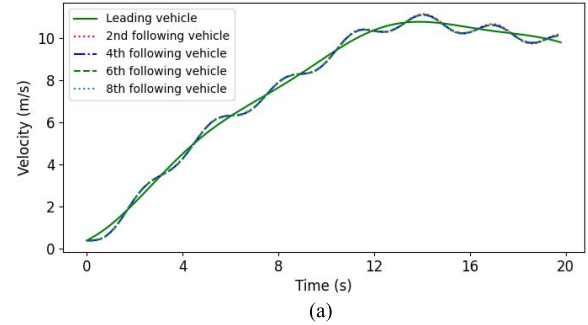

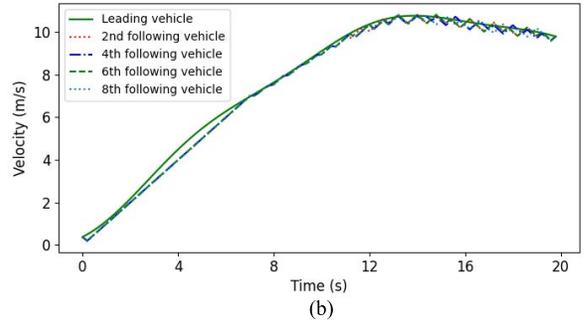

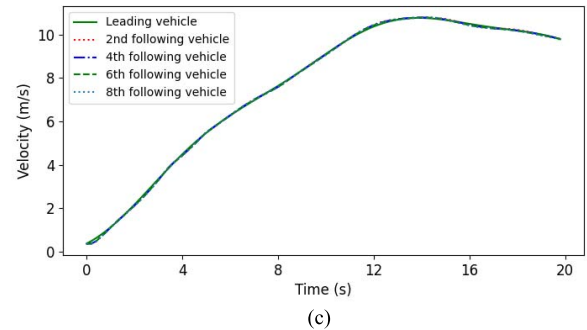

Fig. 4. Velocity of leading and following vehicles in case 1. (a) CACC. (b) DDPG. (c) HCFS.

Fig. 5 shows the jerk of second, fourth, sixth, and eighth following vehicles. All jerks are smaller than $2a_{\max}/\Delta T = 30$, which has been proven by the theorem proposed in Section III-C. According to Table IV, the sum of jerk through HCFS (192.19 m/s$^3$) is much smaller than that of CACC (720.06 m/s$^3$) and DDPG (2344.60 m/s$^3$), and the standard deviation of jerk through HCFS is similar, which implies that it is more comfortable through HCFS in case 1.

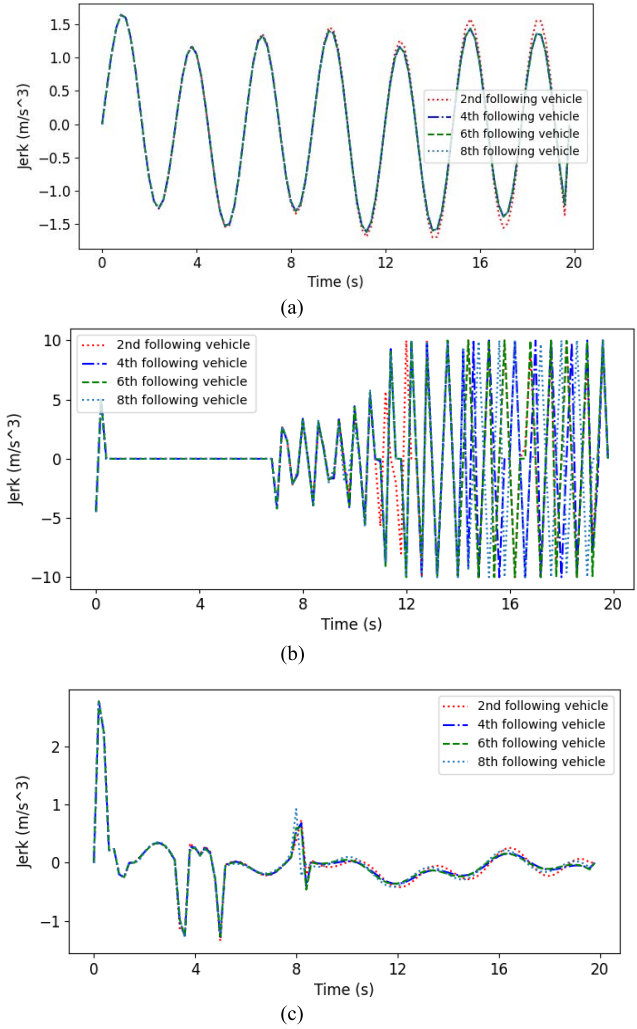

Fig. 5. Jerk of following vehicles in case 1. (a) CACC. (b) DDPG. (c) HCFS.

TABLE IV
RESULTS COMPARISON IN CASE 1

| Item | CACC | DDPG | HCFS |
|---|---|---|---|
| $\sum_{k=1}^{8}\sum_{t=1}^{100} r_k(t)$ | -59.70 | -56.81 | -9.46 |
| $\sum_{k=1}^{8}\sum_{t=1}^{100} |e_{v\_k\_0}(t)|$ (m/s) | 159.39 | 146.22 | 25.26 |
| $\sum_{k=1}^{8}\sum_{t=1}^{100} |jerk_k(t)|$ (m/s$^3$) | 720.06 | 2344.60 | 192.19 |
| $\sqrt{\sum_{k=1}^{8}\sum_{t=1}^{100}(e_{v\_k\_0}(t)-\overline{e}_{v\_k\_0})^2 / N_{case1}}$ (m/s) | 0.22 | 0.16 | 0.04 |
| $\sqrt{\sum_{k=1}^{8}\sum_{t=1}^{100}(jerk_k(t)-\overline{jerk})^2 / N_{case1}}$ (m/s$^3$) | 1.02 | 5.00 | 0.47 |

Fig. 6 shows the switching between CACC and DDPG of the second, fourth, sixth, and eighth following vehicles. The switching is very frequent before 10 s, while CACC is mainly used after 10 s. According to Table IV, the sum of reward of HCFS ($-9.64$) is larger than that of CACC ($-59.70$) and DDPG ($-56.81$). It is the switching mechanism proposed in this study that leads to the performance improvement of HCFS, while the designed rule is used to guarantee that the change rate of acceleration satisfies the constraint during the whole state space in case 1.

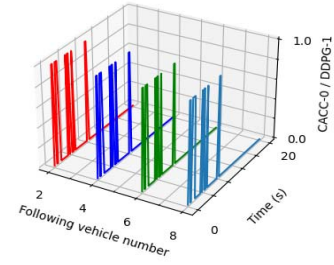

Fig. 6. Switching between CACC and DDPG of following vehicles in case 1.

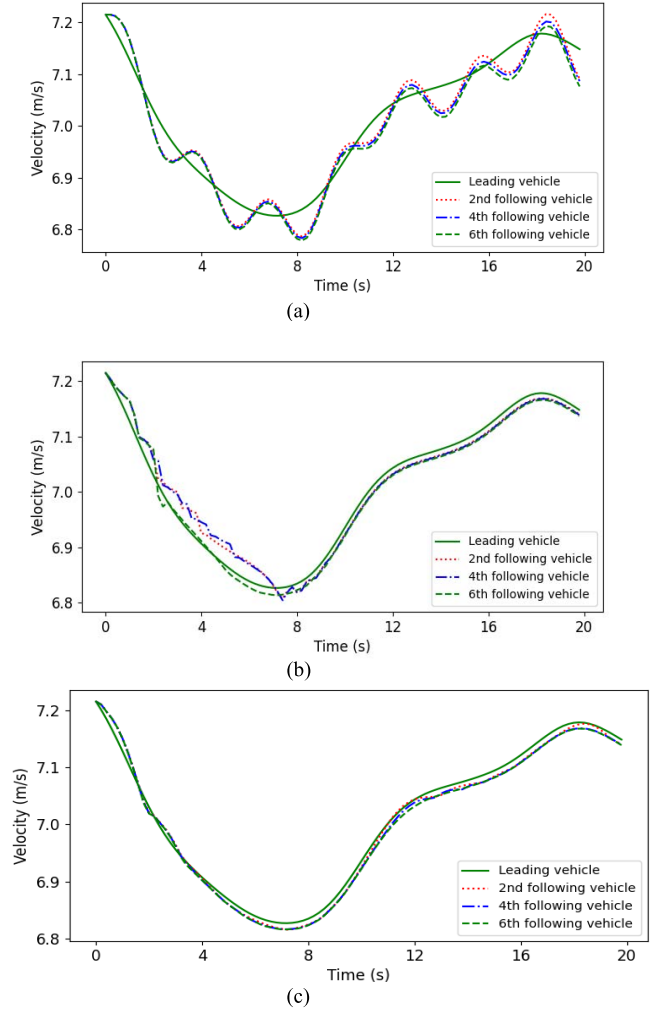

Fig. 7. Velocity of leading and following vehicles in case 2. (a) CACC. (b) DDPG. (c) HCFS.

*B. Case 2*

In this case, one leading vehicle and six following vehicles are considered. The velocity between 620 and 640 s in Fig. 3 is chosen as the velocity of leading vehicle, which is the testing data in case 2. Similar to case 1, the time delay between two adjacent vehicles through V2V is also 5 ms.

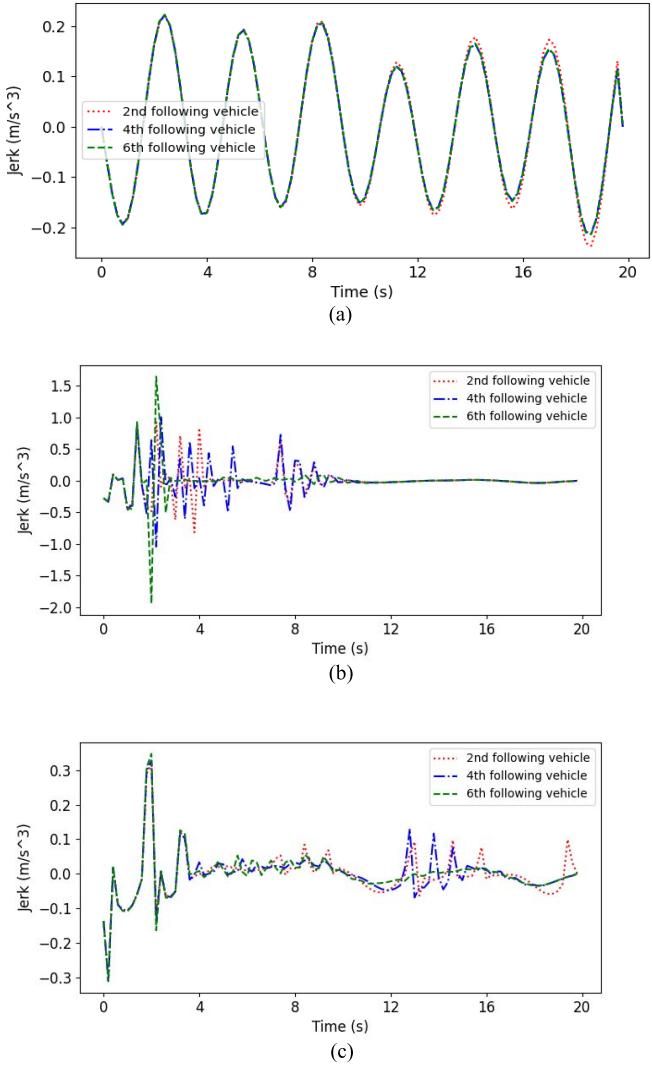

Fig. 8. Jerk of following vehicles in case 2. (a) CACC. (b) DDPG. (c) HCFS.

Fig. 7 shows the velocity of leading vehicle and second, fourth, and sixth following vehicles. As can be seen from Fig. 7(b), the velocity of following vehicles using DDPG deviates from that of the leading vehicle before 8 s, and then, it becomes smooth. To reduce the velocity error with the leading vehicle, CACC and DDPG alternate with each other at the beginning, and DDPG is mainly used in the second half, as shown in Fig. 9. Hence, the velocity of following vehicles using HCFS is improved compared with that of CACC and DDPG in Fig. 7. Table VI also confirms this result, where both the sum of the absolute value of velocity error and its standard deviation through HCFS are smaller than those of CACC and DDPG in case 2.

Fig. 8 shows the jerk of second, fourth, and sixth following vehicles. All jerks of following vehicles are smaller than the threshold, which verifies the proof of theorem in Section III-C. Meanwhile, the sum of jerk and its standard deviation through HCFS are smaller than those of CACC and DDPG according to Table V. Thus, HCFS is more comfortable in case 2 due to the switching mechanism proposed in this study. Moreover, the sum of reward through HCFS is larger than that of CACC and DDPG according to Table V. Therefore, it can be concluded that the car-following performance of HCFS is improved significantly compared with that of CACC and DDPG in the whole state space in case 2.

TABLE V
RESULTS COMPARISON IN CASE 2

| Item | CACC | DDPG | HCFS |
|---|---|---|---|
| $\sum_{k=1}^{6}\sum_{t=1}^{100} r_k(t)$ | -5.86 | -3.54 | -2.10 |
| $\sum_{k=1}^{6}\sum_{t=1}^{100} |e_{v\_k\_0}(t)|$ (m/s) | 15.56 | 9.44 | 5.65 |
| $\sum_{k=1}^{6}\sum_{t=1}^{100} |jerk_k(t)|$ (m/s$^3$) | 67.16 | 66.05 | 24.37 |
| $\sqrt{\sum_{k=1}^{6}\sum_{t=1}^{100}(e_{v\_k\_0}(t)-\overline{e}_{v\_k\_0})^2 / N_{case2}}$ (m/s) | 0.03 | 0.02 | 0.01 |
| $\sqrt{\sum_{k=1}^{6}\sum_{t=1}^{100}(jerk_k(t)-\overline{jerk})^2 / N_{case2}}$ (m/s$^3$) | 0.13 | 0.26 | 0.07 |

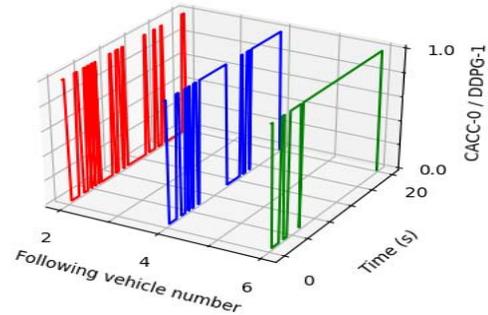

Fig. 9. Switching between CACC and DDPG of following vehicles in case 2.

TABLE VI
RESULTS COMPARISON IN CASE 3

| Item | CACC | DDPG | HCFS |
|---|---|---|---|
| $\sum_{k=1}^{4}\sum_{t=1}^{100} r_k(t)$ | -4.78 | -5.81 | -1.90 |
| $\sum_{k=1}^{4}\sum_{t=1}^{100} |e_{v\_k\_0}(t)|$ (m/s) | 12.83 | 14.72 | 5.04 |
| $\sum_{k=1}^{4}\sum_{t=1}^{100} |jerk_k(t)|$ (m/s$^3$) | 56.69 | 171.48 | 31.60 |
| $\sqrt{\sum_{k=1}^{4}\sum_{t=1}^{100}(e_{v\_k\_0}(t)-\overline{e}_{v\_k\_0})^2 / N_{case3}}$ (m/s) | 0.04 | 0.05 | 0.02 |
| $\sqrt{\sum_{k=1}^{4}\sum_{t=1}^{100}(jerk_k(t)-\overline{jerk})^2 / N_{case3}}$ (m/s$^3$) | 0.17 | 1.18 | 0.12 |

*C. Case 3*

In case 3, one leading vehicle and four following vehicles are considered. The velocity between 1020 and 1040 s in Fig. 3 is chosen as the velocity of leading vehicle, which is the testing data in case 3. The time delay between two adjacent vehicles via V2V is 5 ms.

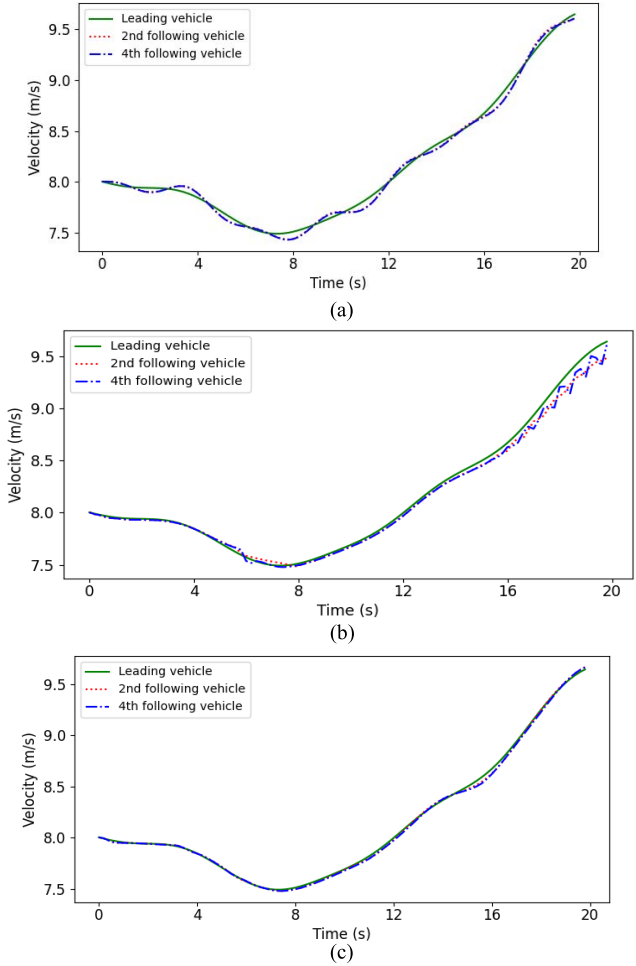

Fig. 10. Velocity of leading and following vehicles in case 3. (a) CACC. (b) DDPG. (c) HCFS.

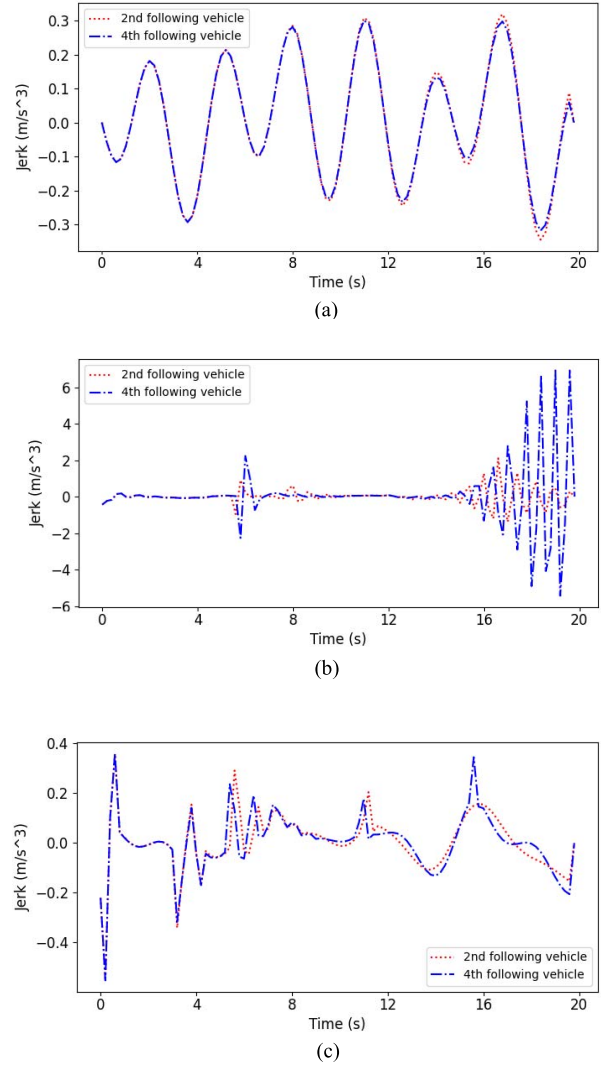

Fig. 11. Jerk of following vehicles in case 3. (a) CACC. (b) DDPG. (c) HCFS.

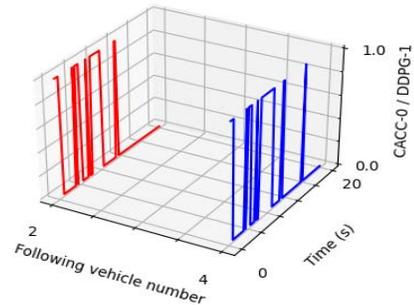

Fig. 12. Switching between CACC and DDPG of following vehicles in case 3.

Fig. 10 shows the velocity of leading vehicle and second and fourth following vehicles using CACC, DDPG, and HCFS, respectively. Similar to Figs. 4(a) and 7(a), the velocity of following vehicles using CACC has periodic oscillation, as shown in Fig. 10(a). As can be seen from Fig. 10(b), the performance of following vehicles using DDPG before 14 s is good, but there is a zigzag oscillation after 14 s. Due to the switching between CACC and DDPG, the velocity of following vehicles using HCFS is almost smooth during the entire process, as shown in Fig. 10(c). This is confirmed by values of velocity errors and the standard deviation in Table VI.

As can be seen from Fig. 11, the jerk of second and fourth following vehicles is smaller than the threshold, where the theorem in Section III-C is further verified by case 3. Similar to cases 1 and 2, the sum of jerk and its standard deviation through HCFS are smaller than those of CACC and DDPG according to Table VI. Moreover, the sum of reward through HCFS ($-1.90$) is larger than that of CACC ($-4.78$) and DDPG ($-5.81$) according to Table VI. Therefore, the results in Figs. 9–11 and Table VI show that the car-following performance through HCFS is better than that of CACC and DDPG in the whole state space in case 3.

## V. CONCLUSION

A type of HCFS is proposed based on DDPG and CACC in this study. On the one hand, CACC is used to guarantee the basic performance of car-following when the performance of DDPG is poor. On the other hand, HCFS makes full use of the exploration ability of DDPG to deal with the car-following cases. Meanwhile, a switching rule is designed to guarantee that the change rate of acceleration is smaller than jerk. Simulation results show that the car-following performance through HCFS is improved significantly compared with that of CACC and DDPG in the whole state space.

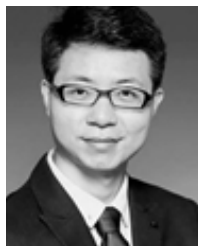

**Ruidong Yan** received the Ph.D. degree from the School of Instrumentation and Optoelectronic Engineering, Beihang University (BUAA), Beijing, China, in 2017.

Then, he did his post-doctoral research at Tsinghua University, Beijing. He is currently a Lecturer with the School of Traffic and Transportation, Beijing Jiaotong University, Beijing. His research interests are intelligent transportation systems, intelligent vehicles, antidisturbance control, and reinforcement learning.

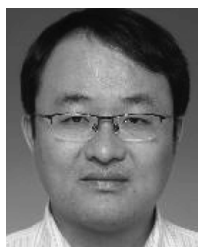

**Rui Jiang** received the B.E. and Ph.D. degrees from the University of Science and Technology of China, Hefei, China, in 1998 and 2003, respectively.

He was an Alexander von Humboldt Research Fellow from 2005 to 2006 and a Japanese Society for Promotion of Science Research Fellow from 2008 to 2009. He is currently a Professor with the School of Traffic and Transportation, Beijing Jiaotong University, Beijing, China. His research interests are traffic flow theory and intelligent transportation systems.

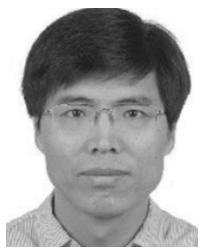

**Bin Jia** received the B.S. and M.S. degrees from Inner Mongolia University of Science and Technology, Baotou, China, in 1997 and 2000, respectively, and the Ph.D. degree in fluid mechanics from the University of Science and Technology of China, Hefei, China, in 2003.

He is currently a Professor with Beijing Jiaotong University, Beijing, China. His research interests are connected automated vehicles, traffic big data mining and application, and traffic flow theory.

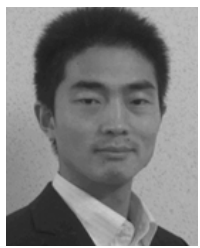

**Jin Huang** received the Ph.D. degree from the College of Mechanical and Vehicle Engineering, Hunan University, Changsha, China, in 2012. He was also a joint Ph.D. in the George W. Woodruff School of Mechanical Engineering, Georgia Institute of Technology, Atlanta, GA, USA, from 2009 to 2011.

He is currently an Associate Research Professor with Tsinghua University, Beijing, China. His research interests are intelligent transportation systems, dynamics control, and fuzzy engineering.

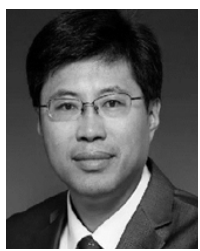

**Diange Yang** received the Ph.D. degree from the School of Vehicle and Mobility, Tsinghua University, Beijing, China, in 2001.

He is currently a Professor with the School of Vehicle and Mobility, Tsinghua University. His research interests are intelligent transportation systems and intelligent vehicles.